\documentclass[sigconf]{acmart}
\renewcommand\footnotetextcopyrightpermission[1]{} 
\usepackage{booktabs} 

\usepackage[symbol]{footmisc}
\setcopyright{none}

\copyrightyear{2018}
\begin{document}
\title{Mammography Assessment using Multi-Scale Deep Classifiers}
\author{Ulzee An}
\authornote{Courant Institute of Mathematical Sciences, New York University}
\affiliation{%
  \institution{New York University}
  \streetaddress{251 Mercer St}
  \city{New York}
  \state{NY}
  \postcode{10012}
}
\email{ua349@nyu.edu}

\author{Khader Shameer}
\authornote{Department of Information Services, Northwell Health}
\authornote{Center for Research Informatics and Innovation, Northwell Health}
\affiliation{%
  \institution{Northwell Health}
  \city{New York}
  \state{NY}
  \postcode{10011}
}
\email{skhader@northwell.edu}

\author{Lakshmi Subramanian $^*$}
\authornote{Center for Data Science, New York University}
\affiliation{%
  \institution{New York University}
  \streetaddress{60 5th Ave}
  \city{New York}
  \state{NY}
  \postcode{10011}
}
\email{lakshmi@cs.nyu.edu}

\begin{abstract}
Applying deep learning methods to mammography assessment has remained a challenging topic. Dense noise with sparse expressions, mega-pixel raw data resolution, lack of diverse examples have all been factors affecting performance. The lack of pixel-level ground truths have especially limited segmentation methods in pushing beyond approximately bounding regions. We propose a classification approach grounded in high performance tissue assessment as an alternative to all-in-one localization and assessment models that is also capable of pinpointing the causal pixels. First, the objective of the mammography assessment task is formalized in the context of local tissue classifiers. Then, the accuracy of a convolutional neural net is evaluated on classifying patches of tissue with suspicious findings at varying scales, where highest obtained AUC is above $0.9$. The local evaluations of one such expert tissue classifier is used to augment the results of a heatmap regression model and additionally recover the exact causal regions at high resolution as a saliency image suitable for clinical settings.
\end{abstract}

\maketitle

\section{Introduction}

  Breast Imaging-Reporting and Data System (BI-RADS) has remained the standard for categorizing the presence of benign and malignant tissue in mammography since its introduction in 1993. To apply it in practice, clinicians undergo years of training to identify minute details indicating various breast conditions which are indicative of cancer to varying degrees. In some cases both local and high-level expressions must be considered. For instance, sickle shaped calcifications are indicative of plasma cell mastitis which is only benign when oriented towards the nipple. \cite{birads} The result of an initial BI-RADS inspection is the basis for bounding outlines which is commonly seen in clinical practice and provided in digital mammography datasets.

The Digital Database for Screening of Mammography (DDSM) has been a popular starting point for applying deep learning to mammography, providing a upwards of 2600 cases with both CC and MLO angle x-rays. As with most clinically annotated tissue data, DDSM also makes available overlays of hand-drawn outlines circumscribing local groupings of suspicious tissue. When observed visually these outlines serve to indicate the general area affected by abnormal growth for visual emphasis, but they are far from pixel accurate labels and only serve as a substitute for ground truth. Unless a model learns the semantics of these boundaries, training directly on the available annotations is unlikely to yield bounds tighter than what is available.

The proposed approach attempts to rectify the issue of loose annotations as well as bridge high accuracy recently demonstrated by breast tissue classifiers \cite{class1, class6} by making them available as alternative source of inference and localization in mammography assessment. An auxiliary classifier is trained to demonstrate expert-level classification accuracy on local tissue samples at varying scales. Then, its pixel-wise gradient which can be used to show saliency is computed at fixed strides covering regions of interest proposed by a heatmap regression model. The saliency of the classifications are treated as strictly tighter bound indicators of suspicious tissue and are aggregated in the original input size, preserving the highest fidelity output resolution possible.

\section{Background and Related Work}

  The need for computer aided diagnosis tools is felt strongly in mammography. The effectiveness of existing procedures is debated with studies showing upwards of $30 \%$ of malignant growths to be retrospectively identifiable in past screenings. \cite{missed, shameer1} Assessment often extend to blind peer screenings since the accuracy and reproducibility of mammography results vary with the expertise of clinicians. Human error in interpretation of medical data is an ongoing topic of study \cite{shameer2} and leveraging machine intelligence is certainly an attractive option.

  While the effectiveness of deep models are well demonstrated in medical imaging, their high parameter count poses an issue in domains where input resolution is mega-pixels in size. Dealing with large data size and general model performance is a broadly studied issue. Classifiers exist which intelligently sample larger data, as proposed by {\it Mnih et al.} \cite{ram} but heuristic sampling runs the risk of missing co-occurring expressions.
 
  The recovery of reasoning information from deep models has also been stressed as a necessary step of validation in the medical domain. Several approaches exist in this regard, including using global average pooling \cite{gaps} to induce visual groupings of inferred causal regions. A more direct approach in saliency verification is to use gradient information with respect to pixels.
  
  In practice, clinical assessment of mammography alternates between phases of broad localization and detailed assessment. Several bodies of work exist on replicating these steps separately as well as proposing an end-to-end deep model. Classification on pre-localized tissue samples have been widely successful \cite{class1} partly by deferring the problem of localization itself. The concept of detecting expressions at multiple scales has also been explored \cite{class2}. Localization itself has been an ongoing area of research. \cite{class4, class5} Among end-to-end models, the importance of preserving native resolution and the necessity for multi-view context is stressed in {\it Geras et al.} \cite{class3} where a classifier views both MLO and CC angles. The saliency visualization method in this study was chosen based on their demonstrated effectiveness in \cite{class3}. {\it Ertosun et al.} proposes a two neural net setup where region proposal only occurs if a preliminary classifier determines the presence of suspicious tissue. \cite{loc1}
  
  The joint segmentation and classification model R-CNN formulated in {\it Girshick et al.} \cite{rcnn} has also been applied to the medical domain with notable success: in brain scan imaging \cite{brainseg} and mammography as well. \cite{seg1} Mammography assessment incur additional challenges for several reasons. X-rays are not cross sections. Three dimensional structures are compressed giving rise to highly noisy expressions. Separation between tumorous and clear tissue can also be a gradual, unlike distinct organs. {\it Ribli et al.} \cite{seg1} demonstrates the highest performance in mammography segmentation.
  
  Deep heatmap regression models \cite{heatmap} have also demonstrated effectiveness in human pose-estimation settings and have not seen wide adoption in the medical image setting. While they are effective for images with guaranteed semantic consistency they require additional filtering layers to learn complex dependencies.

\section{Methods}

We briefly formalize a bottom-up perspective on the relevance of local tissue classifications to an overall assessment. A deep neural net classifier with a softmax layer infers the likelihood of a class given an example $p(y | x)$. In a setting where inferring on full example $x$ is not desirable, it can be substituted by fixed size and fixed stride samples $x_{ij}^\prime$. Each $x_{ij}^\prime$ is used to infer an intermediate assessment $z_{ij}$. With the substitution, the original objective can be approximated as:

\[ p(y|x) \approx \displaystyle \sum_{ij} p(y | z_{ij} \colon C_{ij}) \cdot q(z_{ij} | x^\prime_{ij}) \]

where $C_{ij}$ parameterizes necessary positional context of tissue patch $x^\prime_{ij}$ with respect to full example $x$.

\paragraph{Data Preparation}

All scans were first flipped to one orientation, then cropped without downsampling to 2048-by-2048 pixel regions right-aligned on the nipple. From this base image, tissue samples were generated for testing at three different magnifications: $\times 0.5$, $\times 0.33$ and $\times 0.25$. This procedure was repeated for all available case files in the DDSM database. Left and right scans from individuals were treated as independent samples to increase the variety of available data. Tissue samples for positive and negative cases were generated using separate methods. Findings-positive samples were generated by directly by sampling at the center of mass of annotated regions. Very large annotations were ignored during training to ensure that the subject of the annotation was clearly captured. Negative samples were arbitrarily generated at fixed strides of 32 pixels for regions which did not intersect with any annotations. All tissue samples were treated as independent examples. The content of the samples ranged from empty space, partially visible tissue, and fully covered by tissue.

\paragraph{Tissue Classifier}

Tissue classification has been approached from many different directions. Recent results have shown deep convolutional neural nets obtain state-of-the-art performance for simple lesion identification. \cite{class6, class1} Our tissue-scale expert $q(z_{ij} | x^\prime_{ij})$ was also defined as a convolutional neural net whose architecture has been adopted from VGG, which has demonstrated the ability to classify hundreds of objects. \cite{vgg} The objective of the classifier in our setting was defined as predicting the absence or presence of substantial findings which could be benign or malignant in patches of tissue. This corresponds to BI-RADS assessments of either $\leq$ 1 or $\geq$ 2.


\paragraph{Aggregated Local Results}

To gather local presence information using the trained tissue classifier, it was repeatedly evaluated on the base magnified mammography at moving strides of 64 pixels and to aggregate its local predictions $p(z_{ij}|x^\prime_{ij})$ at offsets $i,j$. The collected predictions at $i,j$ contain condensed global information of suspicious findings to the best of the tissue classifier's ability. 

\begin{figure}
	\centering
  \includegraphics[width=0.33\textwidth]{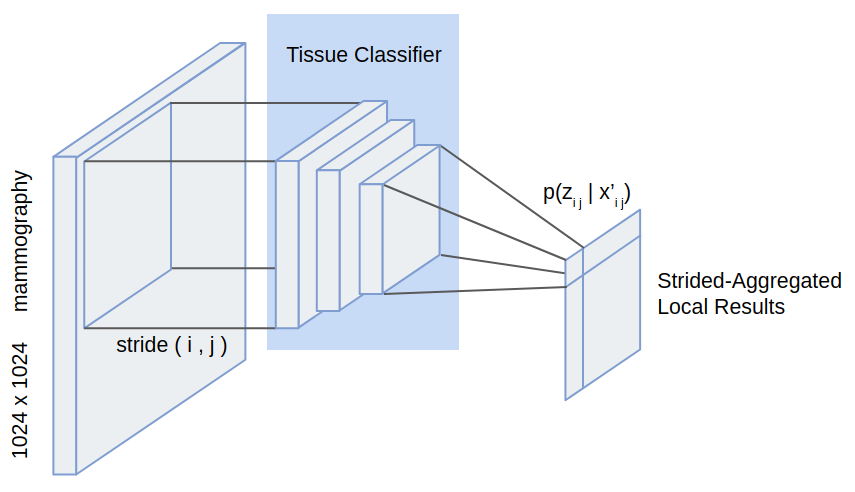}
  \caption{Local result aggregation method at $\times 0.5$ magnification of raw resolution with fixed stride of 64 pixels.}
\end{figure}

\paragraph{Baseline Heatmap Regression}

A deep heatmap regression model \cite{heatmap} was chosen to roughly learn clinical annotations given mammography scans at a downsampled size of 256 pixels. The input image undergoes several layers of convolution with no fully connected layers. This implies heat centers in the output image are direct transformations of expressions found at same location in the input. This inherent property can be advantageous at identifying small structures which must be distinguished from surrounding noise, but is a trade-off against models with fully connected layers which learn complex dependencies. First a baseline version of the heatmap model was tested. Then, a modification was made to introduce the local aggregated results by concatenating them to a final output layer, then adding additional convolution layers to train the regression between the aggregations and the top-level heatmap.

\begin{figure}
	\centering
  \includegraphics[width=0.45\textwidth]{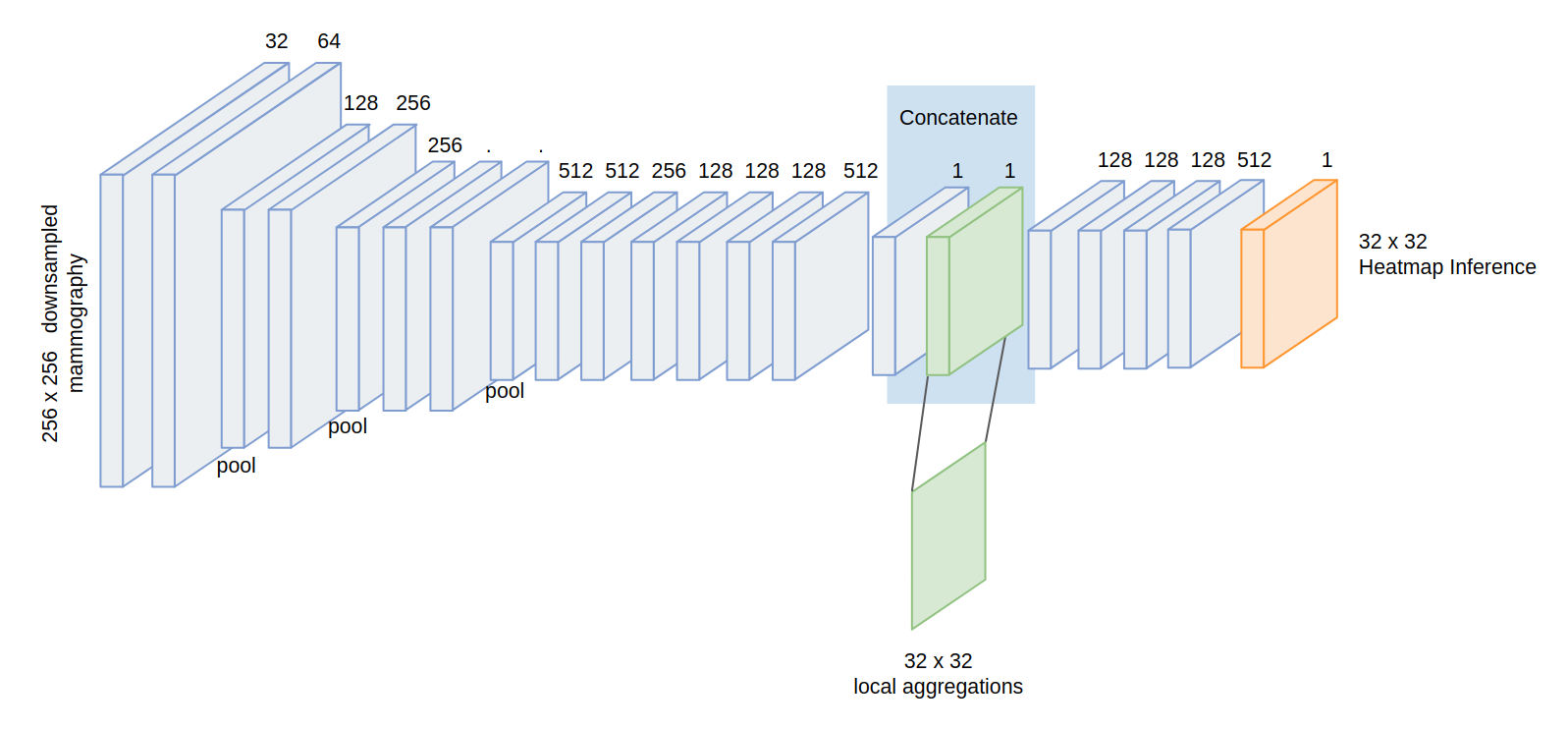}
  \caption{Heatmap regression with auxiliary input of local aggregated results.}
\end{figure}

\paragraph{Pixel-wise Saliency}

The entropy, or confidence, given by prediction $p(y | x)$ is expressed as

\[H(y|x) = -\sum_{y^\prime \in C} p(y^\prime|x) \log{p(y^\prime|x)}\]

The saliency of a given prediction can be visualized by computing the gradient at each pixel with respect to the entropy $|\frac{\delta H_{ij}}{\delta x^\prime_{ij}}|$. \cite{class3} In the absence of $y^\prime$ during inference, the label inferred by the model itself is substituted. At the tissue scale this approach provides pinpoint annotations which are far tighter than circumscribing bounds. During full-scale saliency evaluation, only the saliency of $q(z_{ij}| x^\prime_{ij})$ whose activation $z_{ij}$ appears inside the heatmap inferred by the top-level heatmap regression model are preserved and aggregated.

\section{Results}

\subsection{Classification}

\paragraph{Tissue Classifier}

Of the three classifiers trained at varying magnifications, $\times 0.5$ yielded the best result. This was not surprising given the loss of details with subsequent downsampling. However, as $\times 0.5$ itself was not at max resolution, an acceptable level of downsampling may not impact predictive ability; notably there are visual artifacts from the x-ray digitization process which disappear with downsampling. There were a greater abundance of tissue regions that did not show suspicious expressions, causing an imbalance between positive and negative examples. This was mitigated by applying random brightness, rotations, offsets, flipping, and cropping during training.

Performance of tissue classification approached state-of-the-art and was comparable to results from previous studies. \cite{class1, class6}

Specific types of calcifications are inherently difficult to distinguish from surrounding tissue and would necessarily require both MLO and CC angles to identify. Clinically, this is also the underlying purpose of two standardized angles for mammography. We noted in most failure cases that annotations would be present for lesions that are completely obscured and indistinguishable from surrounding tissue in one angle, but is visible from another angle. It is standard in clinical settings to then annotate the location of the tissue from both angles. The inclusion of such examples in training lowers predictive power in classification under independent treatment of all patches. This semantic trait is resolved in one way and gives strong motivation to a multi-view setup as proposed in {\it Geras et al.} \cite{class3} In the tissue classification case, annotated regions can be paired from both CC and MLO angles, then shown simultaneously to a classifier which learns one set of features using either region.

\begin{table}
  \caption{Data split between stages of evaluation (findings positive / negative).}
  \centering
  \begin{tabular}{lll}
    \toprule
    Stage     & Tissue Patches     & Full Mammography (CC) \\
    \midrule
    Training & 3038 / 311452    & 1432 / 1984     \\
    Validation     & 32 / 32    & 10 / 10      \\
    Testing     & 256 / 256     & 100 / 100  \\
    \bottomrule
  \end{tabular}
\end{table}

\begin{figure}
	\centering
  \includegraphics[width=0.45\textwidth]{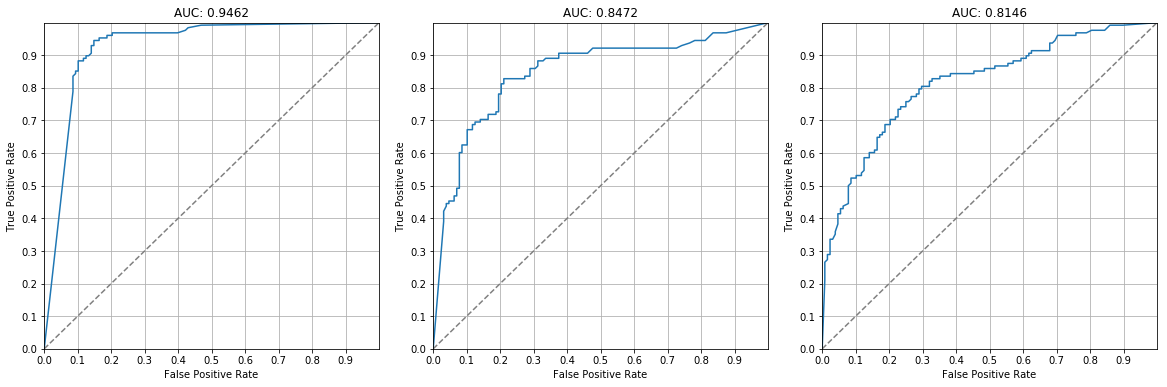}
  \caption{ROC on held-out data of tissue assessment task. Left to right: Classifier trained at $\times 0.5$, $\times 0.33$, and $\times 0.25$ magnifications.}
\end{figure}

\begin{table}
\caption{Breakdown of binary prediction on held-out tissue samples (percentage out of 256 reserved samples).}
  \centering
  \begin{tabular}{lccccc}
    \toprule
    Model & TP & FP & TN & FN & Total Error\\
    \midrule
    
     Tissue $\times 0.5$  & 47.7  & 9.0 & 41.0 & 2.3 & 11.3   \\
     Tissue $\times 0.33$  & 29.7  &  3.9 &  46.0 &  20.3 &  24.2   \\
     Tissue $\times 0.25$  & 33.6  &  6.3 &  43.8 &  16.4 &  22.7   \\
    \bottomrule
  \end{tabular}
\end{table}

\paragraph{Baseline Heatmap Regression}

As defined, the heatmap regression model undergoes max-pooling steps which result in a final heatmap output size of 32-by-32 pixels. The inferred heatmaps were scored using pixel-wise mean squared error (MSE) against downsampled clinical annotations which were Gaussian blurred to reduce the effect of annotation semantics and improve generalizability. With the introduction of locally aggregated results, a small improvement in loss was noted in validation with a larger average loss during training, which indicates a degree of reduced overfitting given the additional information.

The experiment gives motivation to a bottom-up approach for full BI-RADS localization and classification as opposed to a top-down approach which requires high computational resources and the need to sift through a significant amount of noise. Notably there exists a discrepancy between tissue classification accuracy which is expected to be produce above $0.9$ AUC while overall mammography assessment accuracy often falls short except for very specific successful cases such as {\it Ribli et al.} \cite{seg1} The discrepancy is mainly explained by difficulty in localization, but with sufficiently accurate tissue classifier $q(z_{ij} | x^\prime_{ij})$ and informative positional context encoding $C_{ij}$, aggregated classifications should yield the foundations for a highly accurate top-level assessment.

\begin{table}
\caption{Mean-Squared Error between predicted and annotated regions with malignant findings at 256-pixel output resolution.}
  \centering
  \begin{tabular}{lcc}
    \toprule
    Model     & Baseline & $+$ local results ($\times 0.5$)  \\
    \midrule
     Training   & 0.12 & 2.25  \\
    Validation  & 6.48 & 6.04      \\
    \bottomrule
  \end{tabular}
\end{table}

\begin{figure}
	\centering
  \includegraphics[width=0.45\textwidth]{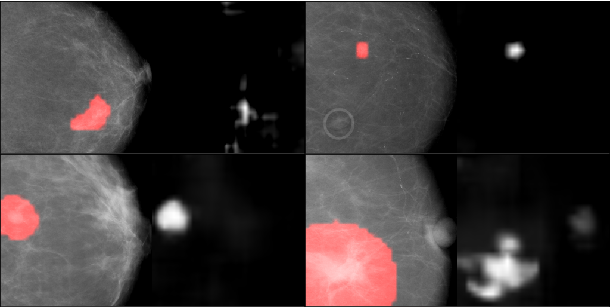}
  \caption{Examples of heatmap inference on held-out mammograms.}
\end{figure}

\subsection{Saliency Retrieval}

Saliency from local tissue classifications were considered in the final visualization only if its activation occurred within corresponding regions inside the final predicted heatmap and if the activation itself indicated a positive prediction. Additionally, the predicted heatmap served as a tight bound and any noise placed by saliency outside its boundaries were ignored. The locally computed saliency images were additively aggregated to their global position in the overall scan at their corresponding magnification level. The overall effect of the procedure highlighted the offending tissue expressions with pinpoint accuracy.

\begin{figure}
	\centering
  \includegraphics[width=0.45\textwidth]{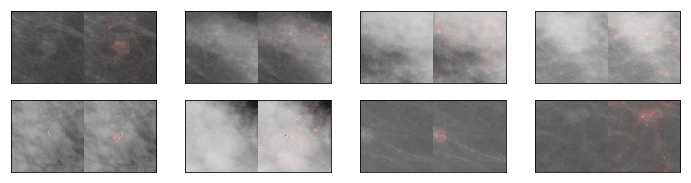}
  \caption{Examples of held-out tissue samples which were correctly classified and their saliency displayed.}
\end{figure}

\begin{figure}
	\centering
  \includegraphics[width=0.33\textwidth]{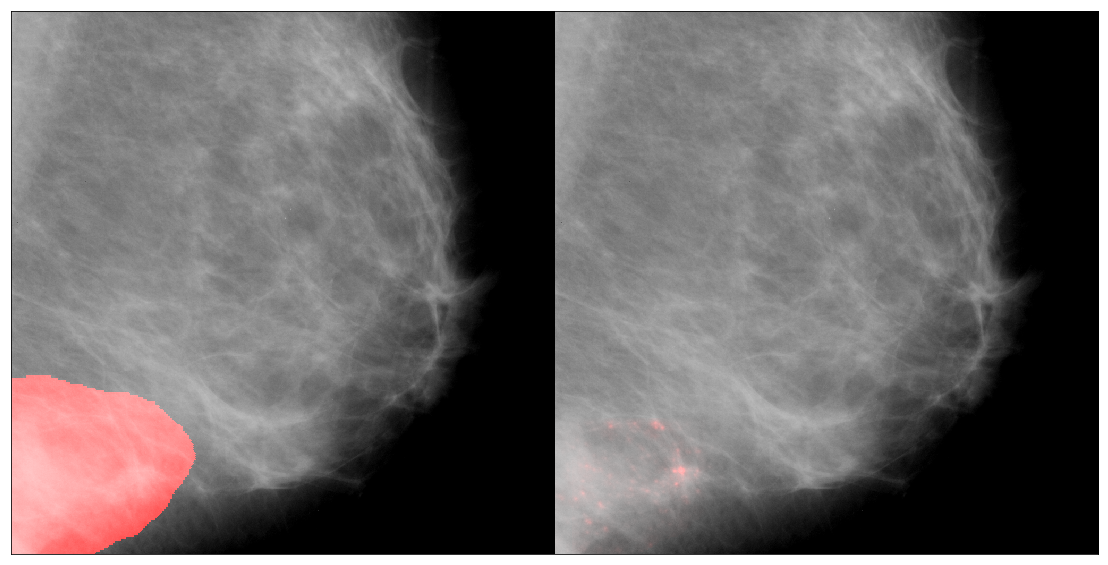}
  \includegraphics[width=0.33\textwidth]{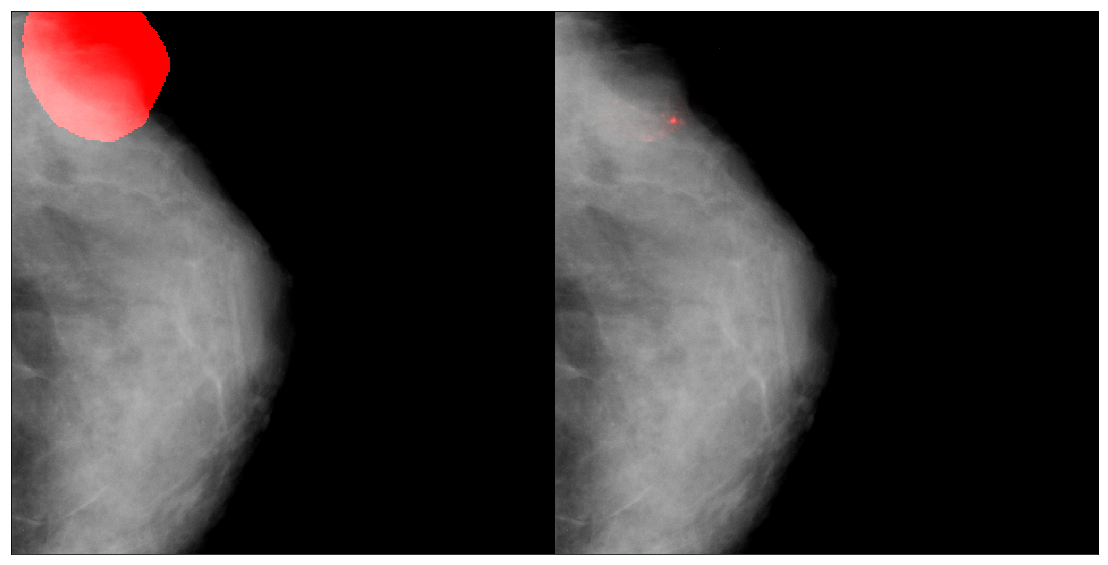}
  \caption{Examples of pixel-level saliency extracted shown in comparison to the original annotations.}
\end{figure}

\section{Conclusion}

Classification tasks on dense tissue images pose several challenges to vanilla deep learning methods. The problem of mammography assessment was broken down into a classification task specializing in discriminating tissue expressions locally then a full context heatmap regression model which guides the aggregation of local results. High accuracy was demonstrated for tissue scale classification and the results of the proposed saliency evaluation method was demonstrated to enhance baseline clinical annotations.

\bibliographystyle{ACM-Reference-Format}
\bibliography{sample-bibliography}

\end{document}